\documentclass[letterpaper, 10 pt, conference]{IEEEconf}

\pdfoutput=1

\IEEEoverridecommandlockouts

\usepackage{cite}
\usepackage{amsmath,amssymb,amsfonts}
\usepackage{algorithmic}
\usepackage{graphicx}
\usepackage{hyperref}
\usepackage{textcomp}
\usepackage{xcolor}
\def\BibTeX{{\rm B\kern-.05em{\sc i\kern-.025em b}\kern-.08em
    T\kern-.1667em\lower.7ex\hbox{E}\kern-.125emX}}

\usepackage{multirow}

\usepackage{bm}
\usepackage[caption=false]{subfig}
\usepackage{siunitx}

\newcommand{\sss}{\scriptscriptstyle}
\DeclareMathOperator*{\argmax}{arg\,max}
\DeclareMathOperator*{\argmin}{arg\,min}

\title{\LARGE \bf 
Factor Graph-Based Continuum Robot State Estimation\\ with Magnus Expansion Priors
}

\author{Lorenzo Ticozzi$^{1}$, Sabyasachi$^{2}$ Dash, Yangkun$^{2}$ Liu, Patricio A.~Vela$^{3}$, and Panagiotis Tsiotras$^{1}$
\thanks{This work has been submitted to the IEEE for possible publication. Copyright may be transferred without notice, after which this version may no longer be accessible.}%
\thanks{*This work has been supported by AFOSR Award FA9550-23-1-0723 and NSF Award FRR-2530577.}
\thanks{$^{1}$Lorenzo Ticozzi and Panagiotis Tsiotras are with the School of Aerospace Engineering, 
        Georgia Institute of Technology, Atlanta (GA), USA
        {\tt\small \{lorenzo, tsiotras\}@gatech.edu}}%
\thanks{$^{2}$Sabyasachi Dash and Yangkun Liu are with the Department of Industrial and Enterprise Systems Engineering, University of Illinois Urbana-Champaign.
        {\tt\small \{dash8, yangkun2\}@illinois.edu}}%
\thanks{$^{3}$Patricio A.~Vela is with the School of Electrical and Computer Engineering, 
        Georgia Institute of Technology, Atlanta (GA), USA
        {\tt\small pvela@gatech.edu}}%
}

\begin{document}

\maketitle

\begin{abstract}
Reconstructing the shape of continuum manipulators from sparse, noisy sensor data is a challenging task, owing to the infinite-dimensional nature of such systems. 
Existing approaches broadly trade off between parametric methods that yield compact state representations but lack probabilistic structure, and Cosserat rod inference on factor graphs, which provides principled uncertainty quantification at the cost of a state dimension that grows with the spatial discretization.
This letter combines the strengths of both these approaches by estimating the coefficients of a compact Geometric Variable Strain (GVS) parameterization within a factor graph framework. 
A novel kinematic factor, derived from the Magnus expansion of the strain field, encodes the closed-form rod geometry as a prior constraint linking the GVS strain coefficients to the backbone pose variables.
The resulting formulation yields a compact state vector, is directly amenable to model-based control, and it retains the modularity, probabilistic treatment, and computational efficiency of factor graph inference. 
The proposed method is evaluated on both simulated and experimental data and compared against a Gaussian Process (GP) regression baseline, demonstrating increased accuracy in partial and sparse measurement scenarios while preserving computational efficiency.
\end{abstract}

\section{Introduction}
Soft robots are increasingly prevalent across diverse application domains, spanning minimally invasive surgery and drug delivery~\cite{cianchetti2018}, robotic-assisted berry picking~\cite{uppalapati2020}, and emerging concepts for on-orbit servicing~\cite{ticozzi2025}.
Interest in soft robots stems from compliant materials that enable safe robot–environment interaction, often with reduced reliance on sensing and active control in contact-rich or unstructured settings~\cite{chen2025}.

When accurate modeling, state estimation, and model-based control are required, however, soft robots present distributed-parameter dynamics with non-trivial geometric nonlinearities that make these tasks substantially more demanding than their rigid-bodied counterparts.
Many works have therefore favored data-driven methods over first-principles models~\cite{thuruthel2017, satheeshbabu2019}, sometimes at the expense of systematic frameworks that generalize across platforms and admit formal analysis.
Motivated by this gap, a strain-based formulation known as Geometric Variable Strain (GVS)~\cite{mathew2025} has led to a reduced-order modeling (ROM) framework enabling efficient derivation of the Lagrangian equations of motion for a broad class of hybrid soft-rigid systems, with earlier Piecewise Constant Strain (PCS)~\cite{renda2016} models recoverable as special cases.

Beyond modeling, planning and feedback control often require an estimate of the soft robot configuration from available measurements---a task commonly referred to as \emph{shape estimation}.
In the GVS context, this amounts to reconstructing the discretized strain coordinates~\cite{talegon2025} and, when needed, their time derivatives, which parameterize the robot configuration.

\subsection{Related Work}

Several shape estimation approaches exist that parameterize the backbone and infer shape parameters from observations.
In~\cite{rosi2022}, SLAM-based camera pose estimation is combined with a fitted Piecewise Constant Curvature (PCC) model~\cite{webster2010}; a similar strategy in~\cite{stella2024} incorporates IMUs to mitigate drift.
In~\cite{loo2019}, an $\mathcal{H}_\infty$-based extended Kalman filter (EKF) estimates PCC generalized coordinates from a Lagrangian dynamic model.
These approaches have been tested only on kinematic PCC representations, which reconstruct backbone pose along the arclength but do not directly recover the underlying strain field.
A GVS-based dynamic observer using a state-dependent Riccati equation (SDRE) is proposed in~\cite{talegon2025}, with strain coefficients as state and actuator readings as measurements.
However, the per-step SDRE cost scales poorly with discretization order, and a growing state dimension can undermine observer well-posedness.

Alternatively, pose and strain fields can be modeled as continuous functions of arclength.
In~\cite{lilge2022}, Gaussian process (GP) regression over $\mathrm{SE}(3)$ estimates both pose and strain along a Cosserat rod, and is extended in~\cite{ferguson2024} to additionally infer external loads.
While this functional formulation provides uncertainty quantification and sparse computational structure, the robot configuration is encoded by a GP posterior rather than a compact state vector, which may hinder downstream model-based control.
More recently,~\cite{ferguson2026} discretizes the full Cosserat rod equations within a factor graph, jointly estimating pose, internal stress, and actuation inputs at each node, hence growing the state dimension with the number of discretization nodes.

A persistent trade-off thus emerges: parametric methods~\cite{rosi2022,stella2024,loo2019,talegon2025} yield compact representations amenable to model-based control, yet lack the computational efficiency and measurement flexibility of graphical-based formulations, while Cosserat rod inference on factor graphs~\cite{lilge2022,ferguson2024,ferguson2026} offers principled uncertainty quantification and sensing modularity, however, at the cost of a state dimension that scales with spatial discretization rather than deformation complexity.

\subsection{Contributions}

This letter bridges the above two paradigms.
Adopting the factor graph formulation of~\cite{lilge2022,ferguson2024,ferguson2026}, we retain its graphical structure and probabilistic treatment, but rather than inferring a continuous strain field or estimating the full rod state at discrete nodes, we estimate the coefficients of a compact GVS parameterization.
To this end, we introduce a novel Magnus factor that encodes the closed-form continuum robot kinematics via the Magnus expansion~\cite{magnus1954, mathew2025} as prior knowledge within the graph.
The proposed method offers three advantages over existing kinematics-based approaches: 1) improved accuracy, 2) robustness to sparse measurement scenarios, and 3) a compact strain representation directly amenable to model-based control design.

\section{Mathematical Preliminaries}

\subsection{Continuum Robot Kinematics}

\begin{figure}[t]
\centerline{\includegraphics[width=\linewidth]{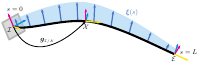}}
\caption{Schematic of a continuum manipulator.}
\label{fig:soft-robot}
\end{figure}

The configuration of a continuum manipulator is described by a curve $\bm{g}_{\sss\mathcal{I}/\mathcal{X}}(\cdot) : [0, L] \to \mathrm{SE}(3)$, parameterized by the arclength coordinate $s \in [0,L]$, where $L$ is the backbone length, $\mathcal{I}$ is a fixed reference frame, and $\mathcal{X}(s)$ is a cross-sectional frame attached to the backbone (Fig.~\ref{fig:soft-robot}), with $\bm{g}_{\sss\mathcal{I}/\mathcal{X}}(0) = \bm{I}_4$ and end-effector frame $\mathcal{E} \triangleq \mathcal{X}(L)$.

The evolution of the cross-sectional pose along the arclength is governed by the following kinematic equation~\cite{renda2017},
\begin{align}
\label{eq:soft-kin}
    \bm{g}^\prime(s) = \bm{g}(s)\bm{\xi}(s),
\end{align}
where $(\cdot)^\prime \triangleq \partial(\cdot)/\partial s$, subject to the initial condition $\bm{g}(0) = \bm{I}_4$.
Here, $\bm{\xi}(s) \in \mathfrak{se}(3)$ is the strain field, with angular and linear strain components denoted by $\bm{k}(s) \in \mathbb{R}^3$ and $\bm{p}(s) \in \mathbb{R}^3$, respectively, with $\bm{k}=[k_x, k_y, k_z]^\top$ and $\bm{p}=[p_x, p_y, p_z]^\top$.

The solution to Eq.~\eqref{eq:soft-kin} admits the closed-form expression
\begin{equation}
    \label{eq:g_prime_sol}
    \bm{g}(s) = \exp{\big(\bm{\Omega}(s)\big)},
\end{equation}
where $\bm{\Omega}(s) \in \mathfrak{se}(3)$ is a convergent infinite series known as the Magnus expansion, obtained via nested commutators of the integral of the strain field with its integral~\cite{magnus1954},
\begin{equation}
    \label{eq:Mag_exp_strain}
    \bm{\Omega}(s) = 
    \int_0^s \bm{\xi}(\sigma) \mathrm{d}\sigma 
    + \frac{1}{2}\int_0^s \left[\int_0^{\sigma}\bm{\xi}(\tau)\mathrm{d}\tau,\,
    \bm{\xi}(\sigma) \right] \mathrm{d}\sigma + \cdots
\end{equation}
In this work, for implementation purposes, the Magnus series in Eq.~\eqref{eq:Mag_exp_strain} is approximated via the same fourth-order quadrature scheme employed in~\cite{mathew2025}.

To render the problem finite-dimensional, we adopt the Geometric Variable Strain (GVS) parameterization~\cite{mathew2025}, where the strain field $\bm{\xi}(s)$ is approximated by a projection onto a finite set of basis functions $\bm{\Phi}_{\bm{\xi}}$,
\begin{equation}
    \label{eq:linear_strain_param}
    \bm{\xi}(s ; \bm{q}) = \bm{\Phi}_{\bm{\xi}}(s)\bm{q} + \bm{\xi}^*.
\end{equation}
In Eq.~\eqref{eq:linear_strain_param}, $\bm{q} \in \mathbb{R}^S$ is a vector of generalized strain coordinates, $\bm{\xi}^* \in \mathfrak{se}(3)$, with coordinate vector $[0,0,0,0,0,1]^\top$, is the reference strain state corresponding to the straight rod configuration, and the strain bases matrix $\bm{\Phi}_{\bm{\xi}} \in \mathbb{R}^{6\times S}$ can encode different basis types, such as monomials, Legendre, B-splines, etc.
Substituting Eq.~\eqref{eq:linear_strain_param} into Eq.~\eqref{eq:Mag_exp_strain} and applying Eq.~\eqref{eq:g_prime_sol}, the full configuration of the soft robot is expressed compactly in terms of the $S$-dimensional vector $\bm{q}$.

\subsection{Factor Graphs}

A factor graph is a bipartite graphical model that represents the factored structure of a joint posterior~\cite{dellaert2017}. 
Given a state $X_k$ and observations $Z_k$, Bayes' rule yields
$p(X_k | Z_k) \propto p(X_k)\, p(Z_k | X_k)$, and the maximum a posteriori (MAP) estimate is
\begin{equation}
\label{eq:map1}
     X^{\mathrm{MAP}}_k = \argmax_{X_k}\, p(X_k)\, p(Z_k | X_k).
\end{equation}
When each factor is modeled as a Gaussian potential, the MAP estimate reduces to a nonlinear least-squares problem that can be solved efficiently; see~\cite{dellaert2017} for a comprehensive treatment on the subject.

\section{Problem Formulation}

\label{sec:prob-form}
We consider the problem of simultaneously estimating the pose field $\bm{g}_{\sss\mathcal{I}/\mathcal{X}}(s)$ and the strain field $\bm{\xi}(s)$ of a continuum manipulator from a set of discrete, sparse, noisy measurements. 
Since the strain is parameterized as in Eq.~\eqref{eq:linear_strain_param}, this reduces to a finite-dimensional parameter estimation problem.

We define a discrete set of backbone poses $\bm{g}_i \triangleq \bm{g}_{\sss\mathcal{I}/\mathcal{X}}(s_i)$ evaluated at arclength nodes $s_i, i=0,\ldots,N$, where $s_0=0$ and $s_N=L$ denote the root and tip of the manipulator, respectively. 
For simplicity, the nodes are assumed to be uniformly spaced along the backbone, yielding a fixed arclength interval $h = L/N$, though this assumption is not essential to the formulation.

\subsection{Factor Graph Construction}

\begin{figure}[t]
\centerline{\includegraphics[width=\linewidth]{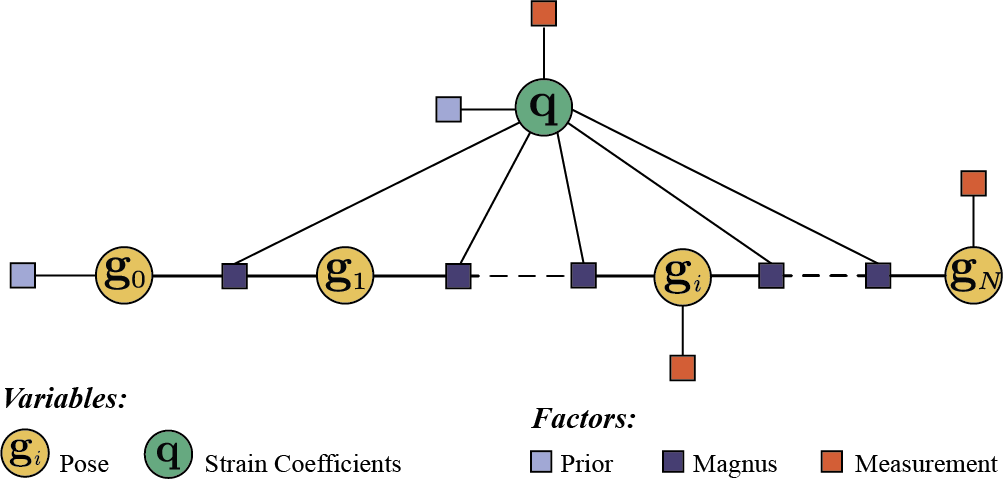}}
\caption{Diagram of the proposed shape estimation factor graph, where $\bm{q}$ collects the strain coefficients in the GVS parameterization, and $\bm{g}_i$ denotes the pose of the $i$-th backbone node.}
\label{fig:soft-robot-graph}
\end{figure}
Within the factor graph framework, the unknown pose variables $\bm{g}_i \in \mathrm{SE}(3)$ and the unknown strain coefficients vector $\bm{q} \in \mathbb{R}^S$ constitute the variable nodes of the graph.
The solution to the shape estimation problem corresponds to the MAP estimate defined in Eq.~\eqref{eq:map1}, specialized as
\begin{equation}
\label{eq:map2}
    \{\bm{g}_0, \ldots, \bm{g}_N, \bm{q}\}^{\mathrm{MAP}} = 
    \argmax_{\bm{g}_0, \ldots, \bm{g}_N, \bm{q}} \, p(\bm{g}_0, \ldots, \bm{g}_N, \bm{q} | \bm{z}),
\end{equation}
where $\bm{z}$ denotes the vector of all available sensor measurements.
To admit a factor graph representation, the joint posterior $p(\bm{g}_0, \ldots, \bm{g}_N, \bm{q} | \bm{z})$ is factored into a product of local potential functions $\phi_i$.
Following standard practice~\cite{dellaert2017}, each factor is modeled as a Gaussian potential $\phi_i \propto \exp{(-\frac{1}{2} \lVert \bm{e}_i \rVert_{\bm{\Sigma}_i}^2)}$, where $\bm{e}_i$ is a residual error vector, $\bm{\Sigma}_i$ its associated covariance matrix, and $\lVert \bm{e}_i \rVert_{\bm{\Sigma}_i}^2=\bm{e}_i^\top \bm{\Sigma}_i^{-1} \bm{e}_i$ the squared Mahalanobis distance.
The resulting factor graph, illustrated in Fig.~\ref{fig:soft-robot-graph}, encodes the full structure of the estimation problem, with variable and factor nodes represented by circles and squares, respectively. 
The construction of each factor class is detailed in the following.

\subsubsection{Magnus Factors}

Recalling Eq.~\eqref{eq:g_prime_sol}, the relationship between two adjacent backbone poses is governed by the underlying strain field, as follows,
\begin{equation}
\label{eq:pose_increment}
    \bm{g}_{i+1} = \bm{g}_i\exp{\big(\bm{\Omega}_i^{i+1}(\bm{q})\big)},
\end{equation}
where $\bm{\Omega}_i^{i+1}(\bm{q})$ denotes the Magnus expansion in Eq.~\eqref{eq:Mag_exp_strain} evaluated over $[s_i, s_{i+1}]$ under the strain parameterization in Eq.~\eqref{eq:linear_strain_param}.
Equation~\eqref{eq:pose_increment} naturally motivates the definition of the following residual vector,
\begin{equation}
    \label{eq:residual-Mag}
    \bm{e}_{i}^M
    \bigl(
    \bm{g}_i,\bm{g}_{i+1},\bm{q}
    \bigr) 
    = 
    \log \Bigl(
    \exp \bigl(
    -\bm{\Omega}_i^{i+1}(\bm{q})
    \bigr)
    \bm{g}^{-1}_i\,\bm{g}_{i+1}
    \Bigr) \in \mathfrak{se}(3).
\end{equation}
The corresponding Magnus factor is then defined as
\begin{equation}
    \label{eq:magnus-fac}
    \phi_i^M 
    \bigl(
    \bm{g}_i,\bm{g}_{i+1},\bm{q}
    \bigr)
    \propto
    \exp{\biggl(-\frac{1}{2} \left\| \bm{e}_{i}^{M}\bigl(
    \bm{g}_i,\bm{g}_{i+1},\bm{q}
    \bigr) \right\|_{\bm{\Sigma}_i^M}^2\biggr)},
\end{equation}
which encodes the kinematic constraint in Eq.~\eqref{eq:pose_increment} over $\bm{g}_i$, $\bm{g}_{i+1}$, $\bm{q}$.

\subsubsection{Measurement Factors}

Measurement factors can be derived for a broad class of sensing modalities, ranging from embedded sensors that measure linear or angular strain along the backbone~\cite{wang2026} to tendon displacement transducers~\cite{talegon2025}. 
In this work, three measurement types are considered: six-dimensional strain measurements, position measurements, and pose measurements.

The strain measurements are modeled as
$
    \tilde{\bm{\xi}}(s) = \bm{\xi}(s) + \bm{w}_{\bm{\xi}}
$,
where $\tilde{\bm{\xi}} \in \mathfrak{se}(3)$ denotes the sensor reading and $\bm{w}_{\bm{\xi}} \sim \mathcal{N}(\bm{0}, \bm{\Sigma}^{\bm{\xi}})$ is zero-mean Gaussian noise.
Given a discrete reading $\tilde{\bm{\xi}}_i = \tilde{\bm{\xi}}(s_i)$ and the strain parameterization in Eq.~\eqref{eq:linear_strain_param}, the following residual is defined,
\begin{equation}
    \label{eq:residual-strain}
    \bm{e}_{i}^{\bm{\xi}}
    \bigl(\bm{q}
    \bigr) =
    \bigl(
    \bm{\Phi}_{\bm{\xi}}^i\bm{q} + \bm{\xi}^*
    \bigr) - \tilde{\bm{\xi}}_i,
\end{equation}
 where $\bm{\Phi}_{\bm{\xi}}^i =\bm{\Phi}_{\bm{\xi}}(s_i)$, yielding the strain measurement factor
\begin{equation}
\label{eq:strain-fac}
    \phi_i^{\bm{\xi}} 
    \bigl(
    \bm{q}
    \bigr)
    \propto
    \exp{\biggl(-\frac{1}{2} \left\| \bm{e}_{i}^{\bm{\xi}}\bigl(
    \bm{q}
    \bigr) \right\|_{\bm{\Sigma}_i^{\bm{\xi}}}^2\biggr)}.
\end{equation}
Notably, $\phi_i^{\bm{\xi}}$ depends solely on the strain coefficient vector $\bm{q}$, since the strain field is fully determined by $\bm{q}$ through Eq.~\eqref{eq:linear_strain_param}.
Pose and position measurement factors, denoted as $\phi_i^{\bm{g}} (\bm{g_i})$ and $\phi_i^{\bm{r}} (\bm{g_i})$, respectively, are obtained analogously by forming the residual between the available measurement and the corresponding pose variable $\bm{g}_i$.

\subsubsection{Prior Factors}

Prior factors are introduced on both $\bm{g}_0$ and $\bm{q}$ to anchor the estimation problem and regularize the strain coefficients, respectively.
For the root pose, the prior enforces the initial condition $\bm{g}_0=\bm{g}(0)=\bm{I}_4$ through the following residual,
\begin{equation}
    \label{eq:g0-residual}
    \bm{e}^p(\bm{g}_0) =
    \log{(\bm{g}_0)} \in \mathfrak{se}(3),
\end{equation}
which measures the deviation of $\bm{g}_0$ from the identity.
For the strain coefficients, centering the prior at the origin yields
\begin{equation}
    \label{eq:q-residual}
    \bm{e}^p(\bm{q}) = \bm{q},
\end{equation}
which is equivalent to Tikhonov regularization on the strain coefficients, with the Tikhonov matrix $\bm{\Gamma}$ chosen such that $\bm{\Gamma}^\top\bm{\Gamma}=(\bm{\Sigma}^p_{\bm{q}})^{-1}$~\cite{hansen1998}.
The corresponding prior densities are
\begin{align}
    \phi^p(\bm{g}_0) &\propto 
    \exp{\biggl(-\frac{1}{2} \left\| \bm{e}^p
    (\bm{g}_0) \right\|_{\bm{\Sigma}^p_{\bm{g}_0}}^2
    \biggr)}, \\
    \phi^p(\bm{q}) &\propto
    \exp{\biggl(-\frac{1}{2} \left\| \bm{e}^p
    (\bm{q}) \right\|_{\bm{\Sigma}^p_{\bm{q}}}^2
    \biggr)},
\end{align}
where $\bm{\Sigma}^p_{\bm{g}_0}$ and $\bm{\Sigma}^p_{\bm{q}}$ encode the tightness of the boundary condition and the degree of regularization imposed on the strain field, respectively.

Given the Magnus, measurement, and prior factors defined thus far, the factor graph in Fig.~\ref{fig:soft-robot-graph} corresponds to the following factorization of the joint posterior,
\begin{equation}
\label{eq:factored-post}
\begin{aligned}
p(\bm{g}_0, \ldots, \bm{g}_N, \bm{q} | \bm{z}) \propto 
\phi^p(\bm{g}_0)\,
\phi^p(\bm{q})
 \prod_{i=0}^{N-1}
\phi_i^M(\bm{g}_i, \bm{g}_{i+1}, \bm{q}) \\
\prod_{j \in \mathcal{S}_{\bm{\xi}}}
\phi_j^{\bm{\xi}}(\bm{q})
\prod_{j \in \mathcal{S}_{\bm{g}}}
\phi_j^{\bm{g}}(\bm{g}_j)
\prod_{j \in \mathcal{S}_{\bm{r}}}
\phi_j^{\bm{r}}(\bm{g}_j),
\end{aligned}
\end{equation}
where $\mathcal{S}_{\bm{\xi}}$, $\mathcal{S}_{\bm{g}}$, $\mathcal{S}_{\bm{r}}$ are the index sets of nodes equipped with strain, pose, and position measurements, respectively.

\subsection{Factor Graph Optimization}
Taking the negative $\log$ of Eq.~\eqref{eq:factored-post} converts Eq.~\eqref{eq:map2} into an equivalent nonlinear least-squares problem (NLLS),
\begin{equation}
\label{eq:nlls}
    \bm{\theta}^{\mathrm{MAP}} = 
    \argmin_{\bm{\theta}} \;
    \left\| \bm{e}(\bm{\theta}) \right\|_{\bm{\Sigma}}^2,
\end{equation}
where $\bm{\theta} \triangleq \{\bm{g}_0, \ldots, \bm{g}_N, \bm{q}\}$, $\bm{e}(\bm{\theta}) \triangleq 
[\bm{e}^p(\bm{g}_0);\, \bm{e}^p(\bm{q});\, \cdots]$, and $\bm{\Sigma} \triangleq \mathrm{blkdiag}(\bm{\Sigma}^p_{\bm{g}_0}, \bm{\Sigma}^p_{\bm{q}}, \ldots)$.
The NLLS in Eq.~\eqref{eq:nlls} can be solved, e.g., via the Levenberg-Marquardt (LM) algorithm.

Note that, in order to enable efficient LM iterations, the analytic Jacobians of each residual in $\bm{e}(\bm{\theta})$ must be supplied to the algorithm.
Specifically, the Jacobians of $\bm{e}_i^M$ with respect to the pose variables $\bm{g}_i$, $\bm{g}_{i+1} \in \mathrm{SE}(3)$ can be computed through standard Lie group perturbation results~\cite{sola2021}.
The strain Jacobian $\partial\bm{e}_i^M/\partial\bm{q}$ follows from the chain rule as
$
    \frac{\partial \bm{e}_i^M}{\partial\bm{q}}=
    \frac{\partial \bm{e}_i^M}{\partial\bm{\Omega}} \frac{\partial \bm{\Omega}}{\partial\bm{q}}, 
$
where the Magnus-specific term $\partial\bm{\Omega}/\partial\bm{q}$ is derived in~\cite{mathew2025}.

\section{Simulations}
\label{sec:simul}
We first evaluated the efficacy of the proposed state estimation approach in simulation. 
Throughout the simulations, we considered a 0.4 m long tendon-driven continuum robot (TDCR) with 14 disks, actuated by three tendons, whose coordinates in the cross-sectional frame are defined by $\boldsymbol{\ell}_i(s)$ for $i = 1, 2, 3$. 
The tendon paths follow $\boldsymbol{\ell}_i(s) = [\ell c_{\theta_i}, \ell s_{\theta_i}, 0]^\top$, with $\ell = 0.01$ m, $c_{\theta_i} \triangleq \cos\theta_i$ and $s_{\theta_i} \triangleq \sin\theta_i$. 
The backbone is modeled as a cylindrical rod with a radius of 1 mm, made of a material with Young's modulus $E = 54$ GPa, Poisson's ratio $\nu = 0.3$, and density $\rho = 6450$ kg/m$^3$. 
The first tendon ($i = 1$) completes one helical path around the backbone, meaning $\theta_1 = 2\pi s / L$, while the remaining two are parallel to the backbone, with $\theta_2 = 2\pi/3$ and $\theta_3 = 4\pi/3$.

\subsection{Data Generation}
Simulation data were generated using the publicly available Cosserat rod solver described in~\cite{rao2021}.
The solver was run for 60 random combinations of tendon tensions, each uniformly distributed between 0~N and 20~N, and an external tip force with three components uniformly distributed between -1.5~N and 1.5~N.
Noisy sensor measurements were generated by injecting noise into the ground-truth simulation data at the specific arclengths corresponding to the sensor locations.
The sensor noise statistics were selected to match those employed in~\cite{lilge2022}, with standard deviations $\sigma_{\omega}=0.01$ rad and $\sigma_{r}=1$ mm for the orientation and position, respectively, and $\sigma_k=\sigma_p=0.05$ for the angular and linear strain components.

\subsection{Choice of Hyperparameters}
The proposed algorithm was implemented using the GTSAM library~\cite{gtsam} to both construct the factor graph and solve the NLLS in Eq.~\eqref{eq:nlls} via the LM algorithm.
All computations were conducted on a 14-core Apple M4 Pro chip.
Since we parameterized the strain field with cubic B-splines, the variable $\bm{q}$ in the factor graph collects the spline coefficients to be estimated.
The robot backbone was discretized into $N=10$ intervals, resulting in 11 pose variables $\bm{g}_0,\ldots,\bm{g}_{10}$.

In all simulations, the Magnus factor covariances were set to $\bm{\Sigma}_i^M=1\cdot10^{-6}\bm{I}_6$ to encode a strong kinematic prior for the estimator.
The LM optimization was carried out in a two-pass continuation scheme, with $\bm{\Sigma}_i^M$ first inflated by 10$^4$ and then restored to its nominal value, to improve robustness to the straight-rod initialization.
The strain covariance $\bm{\Sigma}_{\bm{q}}^p$ was defined as a diagonal matrix, with the elements corresponding to the angular components set to 3000, while the $z$-axis elongation entries were set to 30.
No DOFs were assigned to the shear components $p_x$, $p_y$ throughout the simulations, reflecting the physics of the system, where large deformations are predominantly governed by bending modes rather than shear or elongation.
Finally, the root pose, $\bm{g}_0$, was constrained to the identity matrix using GTSAM's constrained noise model.
The covariance matrices for the measurement factors were assigned consistently with~\cite{lilge2022}, yielding $\bm{\Sigma}_i^{\bm{g}}=\mathrm{diag}([10\sigma_{\omega}^2, 10\sigma_{\omega}^2, 10\sigma_{\omega}^2, 10\sigma_r^2, 10\sigma_r^2, 10\sigma_r^2])$ for the pose measurements, $\bm{\Sigma}_i^{\bm{r}}=10\sigma_r^2\cdot\bm{I}_3$ for the positions, and $\bm{\Sigma}_i^{\bm{\xi}}=\mathrm{diag}([10\sigma_k^2, 10\sigma_k^2, 10\sigma_k^2, 10\sigma_p^2, 10\sigma_p^2, 10\sigma_p^2])$ for the strains.
\begin{figure}[t]
\centerline{\includegraphics[width=\linewidth]{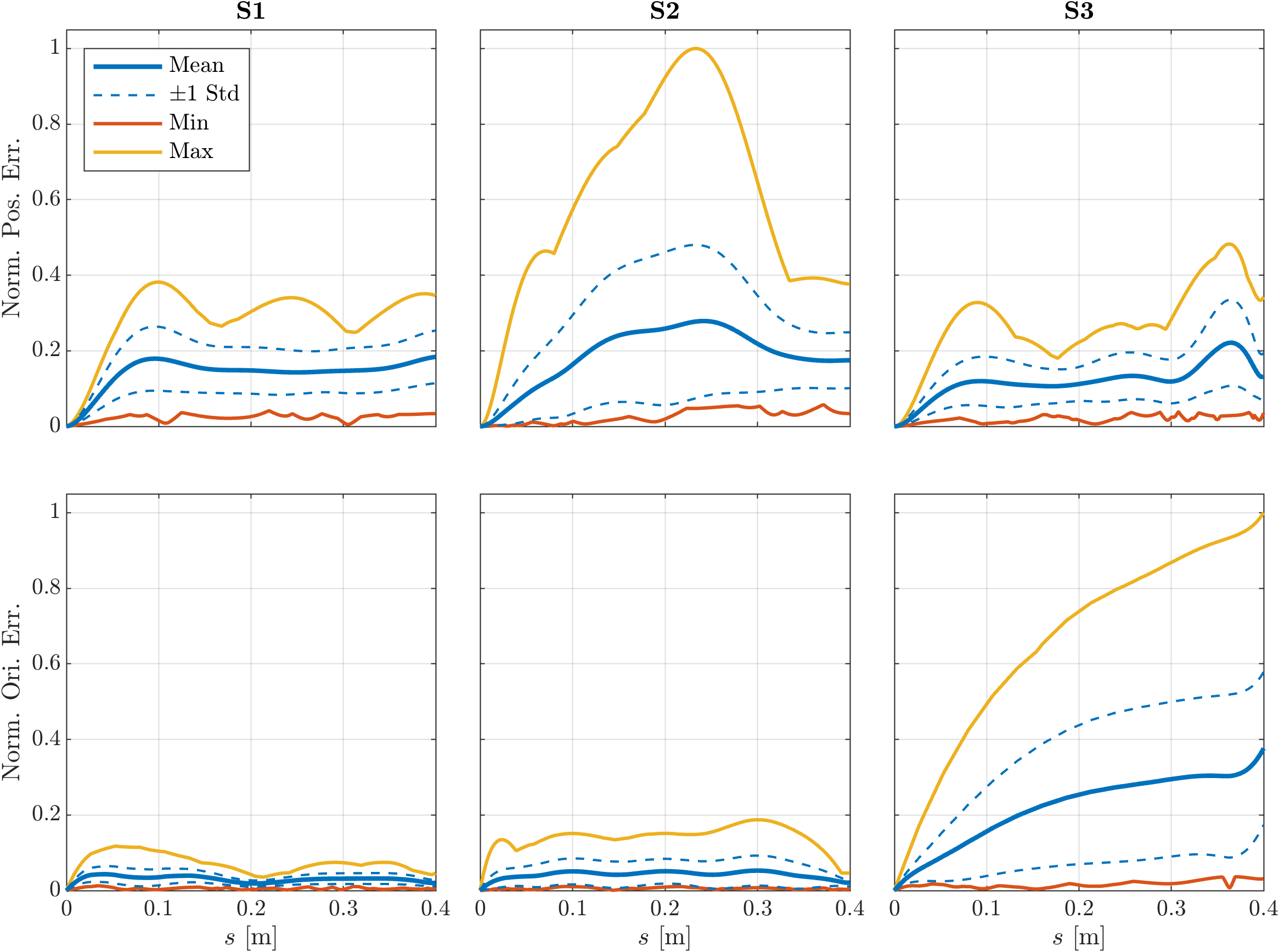}}
\caption{Normalized pose error statistics for scenarios S1, S2 and S3.}
\label{fig:norm-pose-err}
\end{figure}
\begin{table}[b]
\caption{\label{tab:scenarios}
Simulation scenarios ($L=0.4$ m).
}
\centering
\footnotesize
\renewcommand{\arraystretch}{1.2}
\begin{tabular}{@{}llcc@{}}
\hline
\hline
\textbf{ID} & \textbf{Meas.} & \textbf{Arclength Coordinates} & \textbf{Spline Control Points}  \\
\hline
\multirow{2}{*}{S1} & \multirow{2}{*}{Pose} & \multirow{2}{*}{$[L/2,\, L]$} & $n_{k_x}=8$, $n_{k_y}=10$ \\
 & & & $n_{k_z}=5$, $n_{p_z}=5$ \\
\hline
\multirow{2}{*}{S2} & Strain & $[L/4,\, L/2,\, 3L/4]$ & $n_{k_x}=8$, $n_{k_y}=10$ \\
 & Pose & $L$ & $n_{k_z}=5$ \\
\hline
S3 & Position & $[L/5,\, 2L/5,\, 3L/5,\, 4L/5,\, L]$ & $n_{k_x}=4$, $n_{k_y}=4$ \\
\hline
\hline
\end{tabular}
\end{table}
\begin{table}[b]
	\caption{\label{tab:stats}
		Pose error statistics and computational performance for simulation scenarios S1, S2 and S3.
	}
	\centering
	\footnotesize
	\setlength{\tabcolsep}{4pt}
	\renewcommand{\arraystretch}{1.2}
	\begin{tabular}{@{} l c c c c | c @{}}
		\hline
		\hline
		\textbf{ID} &
		\begin{tabular}{@{}c@{}}\textbf{Pos.} \\ \textbf{Error}, mm\end{tabular} &
		\begin{tabular}{@{}c@{}}\textbf{Max. Pos.} \\ \textbf{Error}, mm\end{tabular} &
		\begin{tabular}{@{}c@{}}\textbf{Rot.} \\ \textbf{Error}, deg\end{tabular} &
		\begin{tabular}{@{}c@{}}\textbf{Max. Rot.} \\ \textbf{Error}, deg\end{tabular} &
		\begin{tabular}{@{}c@{}}\textbf{Total} \\ \textbf{Time}, ms\end{tabular} \\
		\hline
		S1 & 1.45$\pm$0.36 & \textbf{3.89} & \textbf{1.46$\pm$0.45} & \textbf{5.64} & \textbf{1.2$\pm$0.31} \\
		\hline
		S2 & 1.92$\pm$1.07 & 10.18 & 2.0$\pm$1.36 & 8.99 & 1.26$\pm$0.5 \\
		\hline
		S3 & \textbf{1.23$\pm$0.32} & 4.91 & 10.60$\pm$7.26 & 48.0 & 1.65$\pm$0.44 \\
		\hline
		\hline
	\end{tabular}
\end{table}

\subsection{Results}
\label{sec:results}
\begin{figure*}[t]
    \centering
    \subfloat[Scenario S1.\label{fig:colors-s1}]{%
        \includegraphics[width=0.32\linewidth]{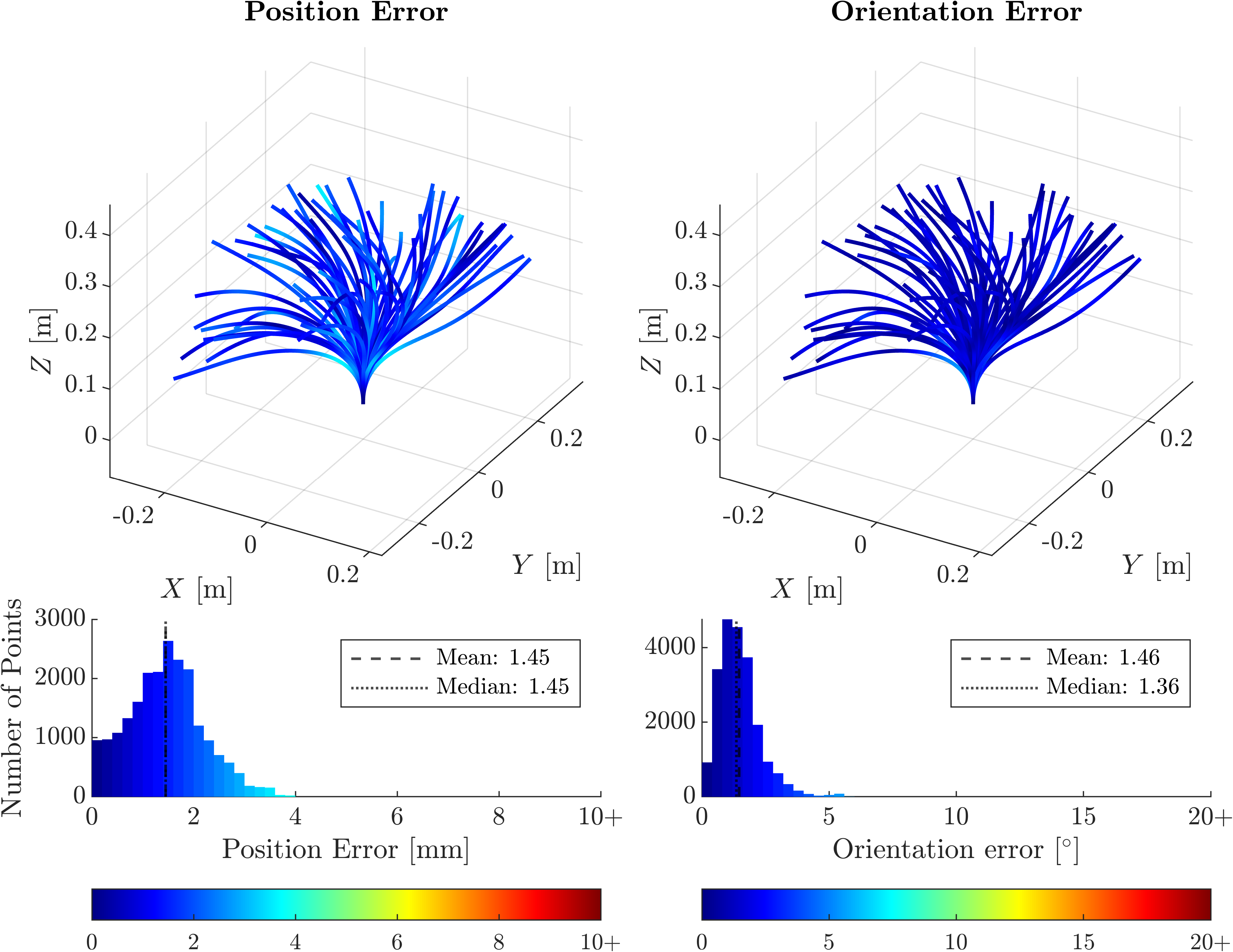}%
    }
    \hfill
    \subfloat[Scenario S2.\label{fig:colors-s2}]{%
        \includegraphics[width=0.32\linewidth]{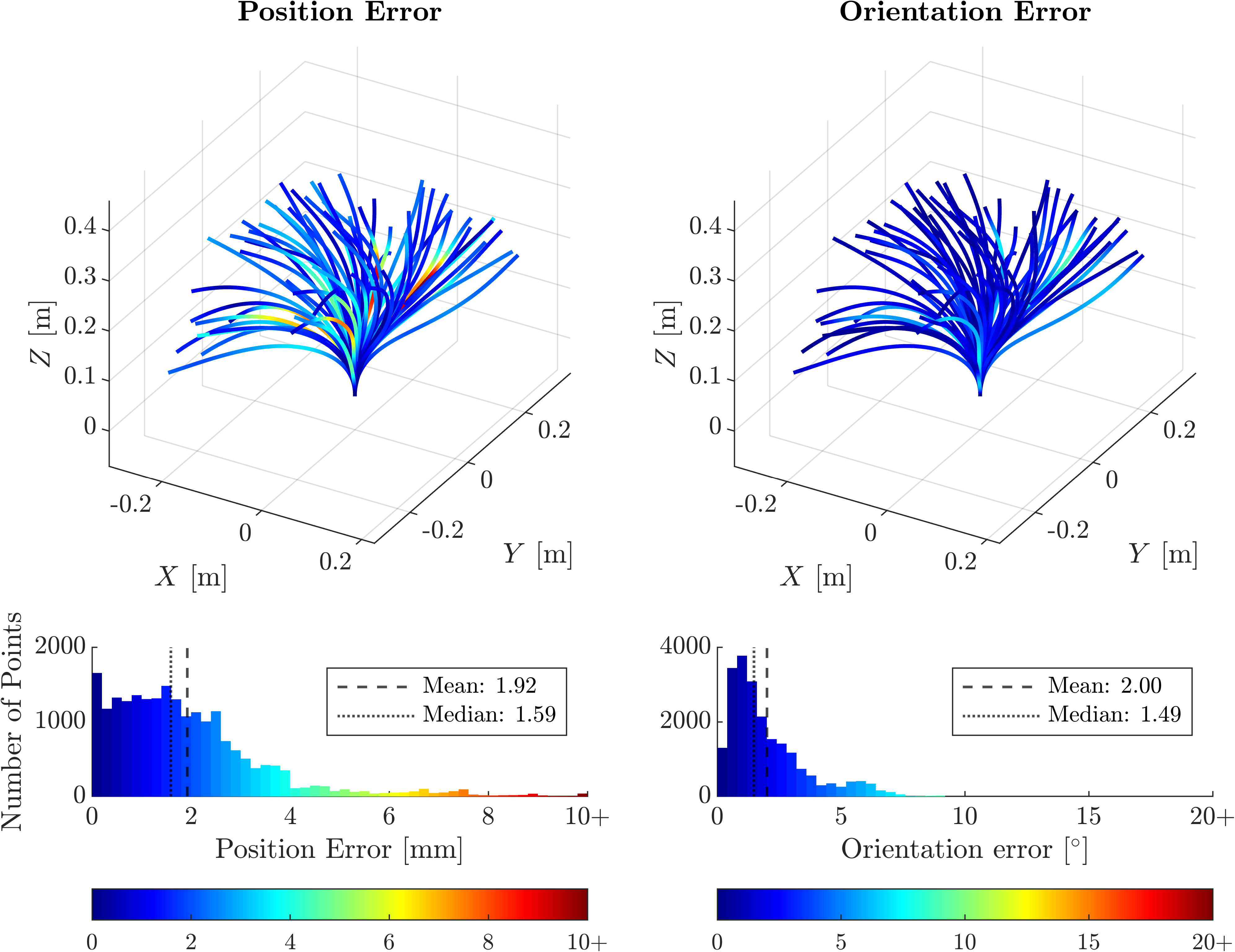}%
    }
    \hfill
    \subfloat[Scenario S3.\label{fig:colors-s3}]{%
        \includegraphics[width=0.32\linewidth]{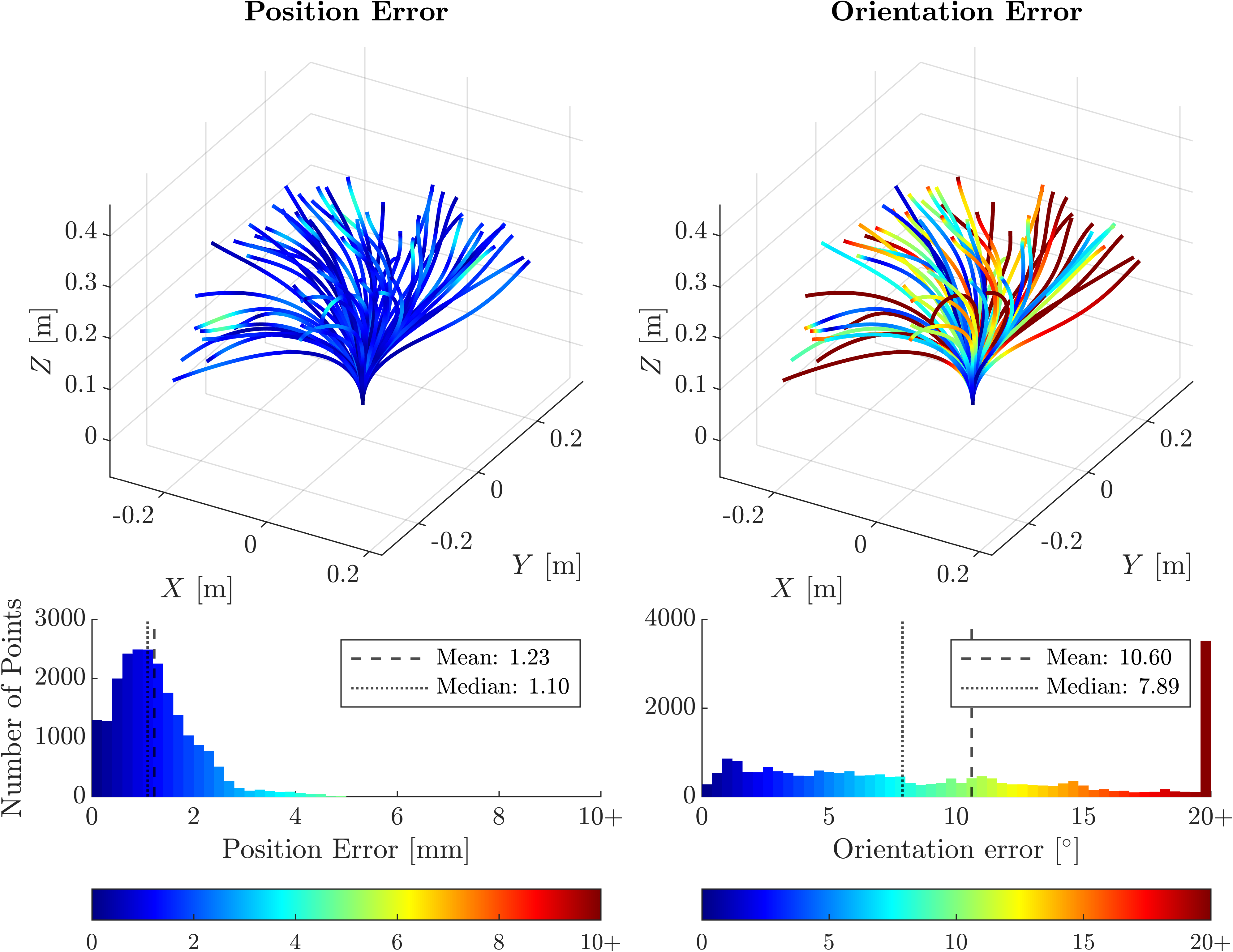}%
    }
    \caption{Reconstructed backbone shapes in simulation scenarios S1, S2 and S3.}
    \label{fig:shape-colors}
\end{figure*}

Three simulation scenarios were considered, denoted S1, S2, and S3, each defined by a specific set of measurements at selected arclength coordinates and a choice of strain basis.
The parameters of each scenario are detailed in Table~\ref{tab:scenarios}, where $n_{k}$ and $n_p$ denote the number of control points associated to the angular and linear strain modes, respectively.
While the backbone pose is only solved for at the estimation nodes, the estimates $\bm{g}_i$ and $\bm{q}$ can be substituted into Eq.~\eqref{eq:pose_increment} to reconstruct the pose at any value of $s$, yielding a continuous backbone estimate that is compared against the ground-truth solutions to assess estimation performance.

Mean pose estimation errors across the 60 ground-truth samples are reported in Fig.~\ref{fig:norm-pose-err} and Table~\ref{tab:stats} (total time includes both optimization and dense query), while the batch shape estimation results with error histograms appear in Fig.~\ref{fig:shape-colors}.
Overall, the best performance is achieved in S1, where pose measurements are available at both the midpoint and the tip.
Scenario S2 shares the same angular strain parameterization as S1 but omits elongation $p_z$ and uses the mixed measurements in Table~\ref{tab:scenarios}, resulting in lower position accuracy and higher variance in the mid-rod region where no position information is available.
In S3, only the $x$- and $y$-axis bending modes are active, and position measurements are provided at five backbone nodes, yielding a mean position error of 1.23~mm.
However, since no direct orientation information is available, orientation errors in S3 are the largest across all scenarios (Figs.~\ref{fig:norm-pose-err}, \ref{fig:colors-s3}).

\subsection{Discussion}

A few aspects are worth discussing to better evaluate the proposed approach.
First, we note that the choice of the strain bases for the GVS parameterization is not unique, but should be driven by several considerations, including the actuator configuration, the number and type of available measurements, and the nature of any external loads.
Our choice of local bases such as B-splines ensures that each basis function has support over only a subset of the arclength domain, so that at any given arclength $s$, matrix $\bm{\Phi}_{\bm{\xi}}$ has at most $p+1$ nonzero entries per row (with $p=3$ for cubic splines).
This induces sparsity in the residual Jacobians and decreases the computational cost of solving the NLLS in Eq.~\eqref{eq:nlls}.
The number of spline control points, which governs the approximation order of each strain component, is also a critical design parameter.
In our simulations, more control points were assigned to the bending modes (see Table~\ref{tab:scenarios}), as these dominate the Cartesian displacement of the manipulator.
A further benefit of encoding the exact rod geometry through the Magnus factors is that, even when only position measurements are available, as in S3, the proposed method infers cross-section orientation with limited error in several cases, as evidenced by Fig.~\ref{fig:colors-s3}; this behavior is further investigated in Sec.~\ref{sec:comparison}.
Finally, the variable node $\bm{q}$ is connected to every pose node along the backbone through the Magnus factors (Fig.~\ref{fig:soft-robot-graph}), reflecting the global dependence of the backbone shape on the GVS strain coefficients.
This dense connectivity reduces the sparsity of the factor graph, partially undermining the computational advantage that sparsity confers to factor graph optimization~\cite{dellaert2017}.
In practice, however, the combination of a low-dimensional strain parameterization and locally supported B-spline bases yields solution times that remain competitive with sparse methods, as highlighted in the next section.

\subsection{Comparison with Baseline Method}
\label{sec:comparison}
\begin{figure}[t]
    \centering
    \subfloat[Scenario S1.\label{fig:comp-s1}]{%
        \includegraphics[width=0.49\linewidth]{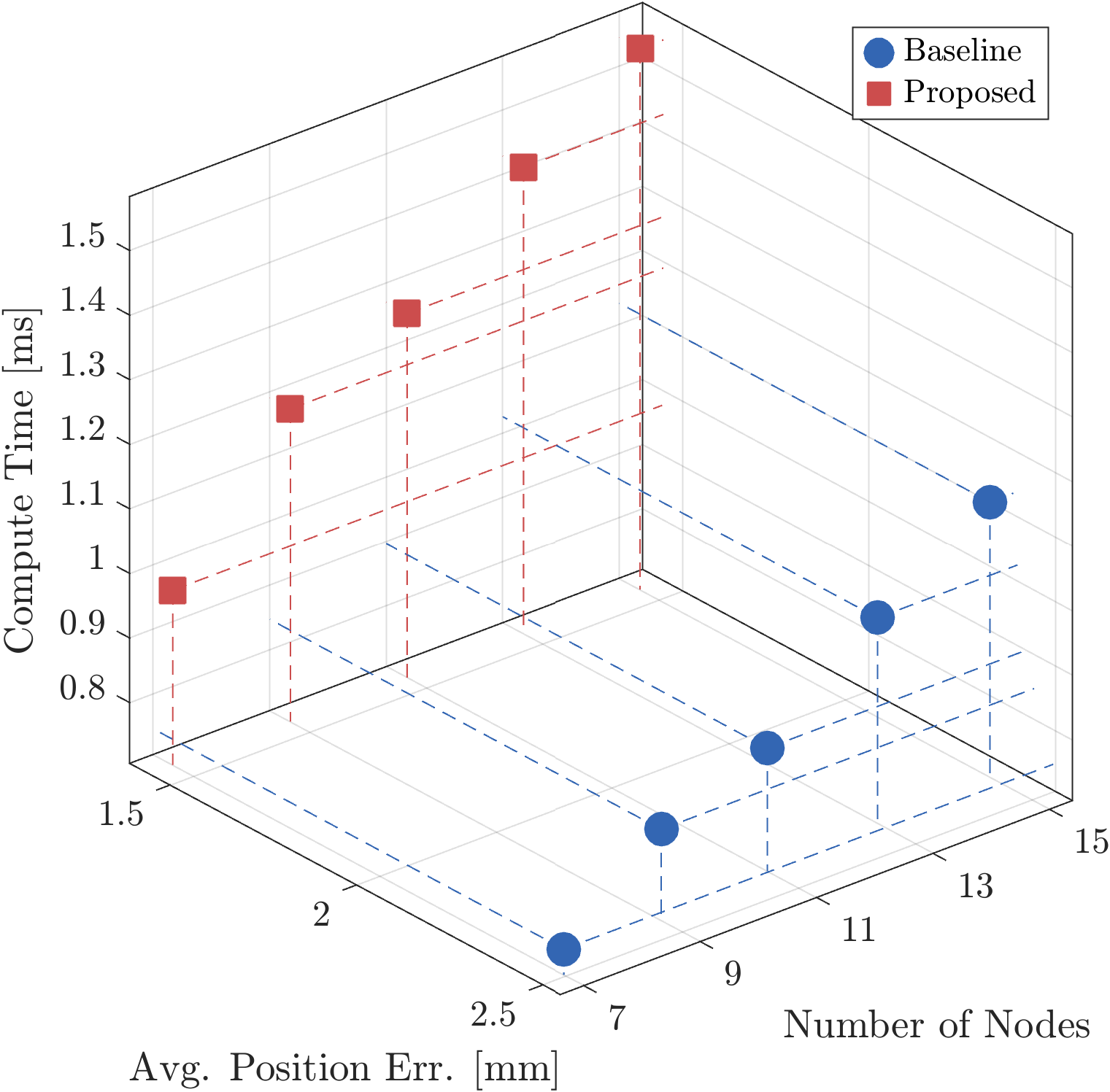}%
    }
    \hfill
    \subfloat[Scenario S2.\label{fig:comp-s2}]{%
        \includegraphics[width=0.49\linewidth]{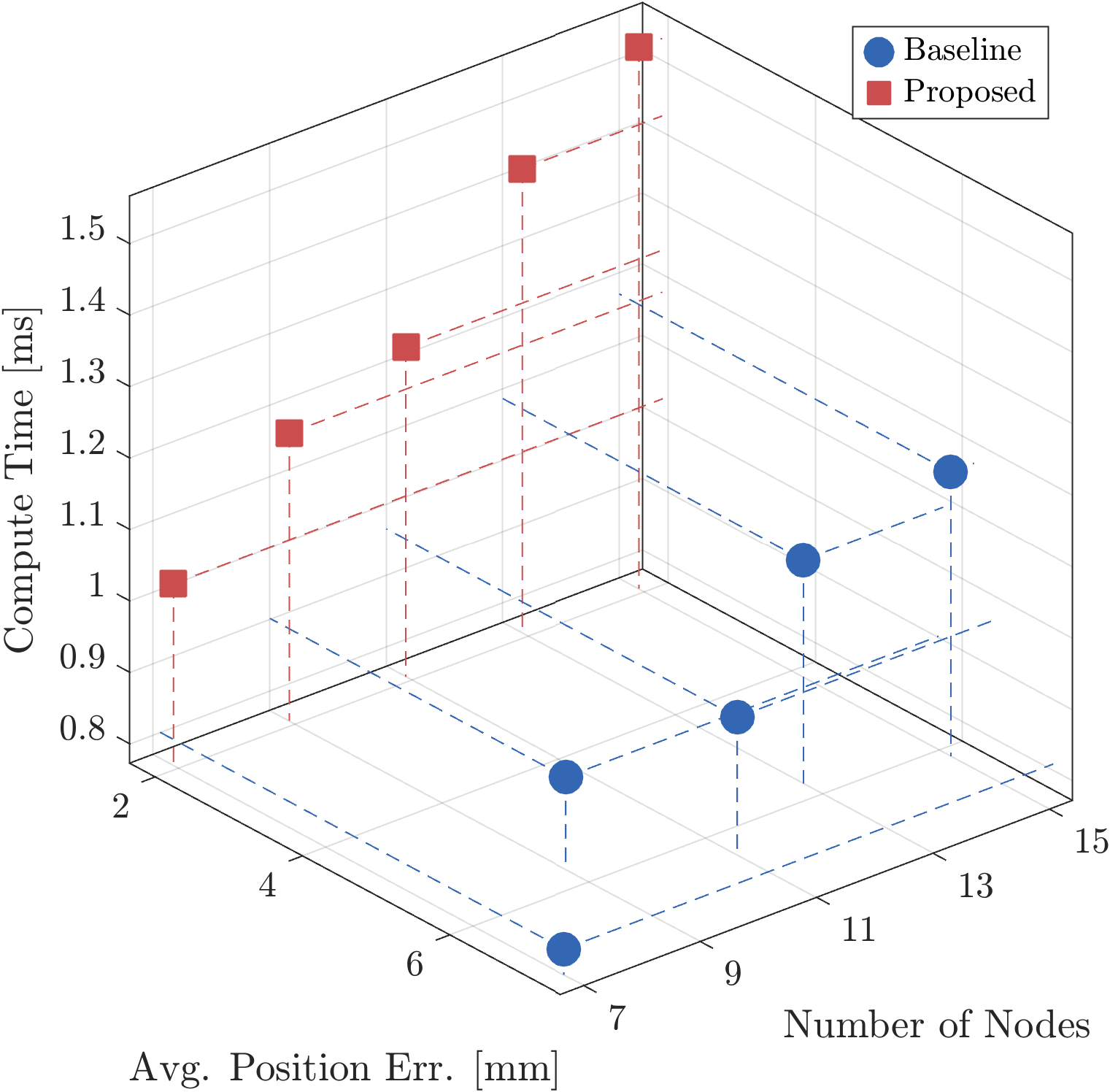}%
    }
    \caption{Accuracy and computation efficiency comparison between the proposed method and the baseline in~\cite{lilge2022}.}
    \label{fig:comparison}
\end{figure}
\begin{figure}[t]
\centerline{\includegraphics[width=\linewidth]{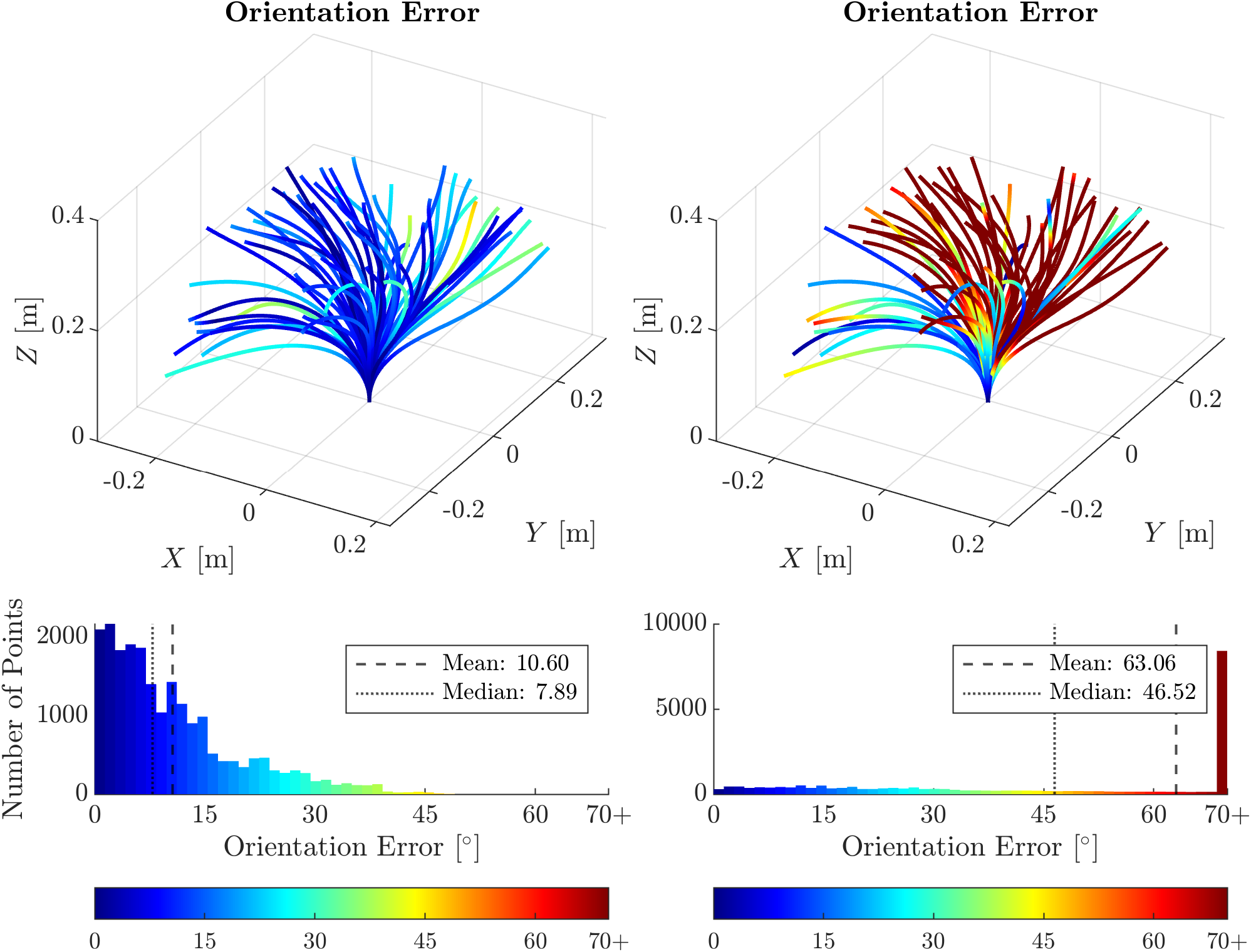}}
\caption{Comparison of the orientation errors in the proposed method (left) and the baseline (right) for the position-only measurements scenario S3.}
\label{fig:comp-ang}
\end{figure}
This section compares the proposed approach against the method described in~\cite{lilge2022} (hereafter ``baseline''), reimplemented in GTSAM to replicate its variables and factors, and using identical hyperparameters wherever applicable (sensor noise statistics, initial conditions, and measurement covariances; see Sec.~\ref{sec:simul}).
Both algorithms were evaluated on scenarios S1, S2, and S3, with mean estimation errors computed across the 60 ground-truth configurations and the backbone length.
For each method, the number of estimation nodes was varied over odd values between 7 and 15 to assess the impact of grid resolution on accuracy and computation time.
For the baseline, solution queries at ground-truth nodes were performed via GP interpolation (see Sec.~2.2.8 in~\cite{lilge2022}).
In the interest of space, only position errors for S1 and S2 are reported in Fig.~\ref{fig:comparison}.

For an equal number of estimation nodes, the proposed method yields higher accuracy across all simulations.
Beyond the accuracy gap, two structural limitations emerge in the baseline.
First, the angular variables remain poorly constrained in the absence of direct orientation measurements, as seen in Fig.~\ref{fig:comp-ang} (right), since the chosen prior factor does not encode the nonlinear coupling between linear and angular strain components.
Second, strain measurements are enforced at the nearest estimation node rather than at their true arclength coordinate, as the baseline has no mechanism to attach factors at continuous locations between nodes.
As the grid spacing changes with the number of estimation nodes, the snapping error varies non-monotonically, reaching zero whenever a true measurement location coincides exactly with a node and growing otherwise, producing the irregular accuracy pattern observed in Fig.~\ref{fig:comp-s2}.

In the proposed method, the strain DOFs are fixed by the number of spline control points independently of the number of nodes $N$, and the Magnus factor residual is evaluated at the exact continuous arclength through the B-spline model, so measurement locations are always honored precisely regardless of grid density.
Furthermore, the commutator term in Eq.~\eqref{eq:Mag_exp_strain} encodes the nonlinear coupling between bending and torsional strain at the Lie algebra level, implicitly constraining the orientation estimates even in the absence of direct orientation measurements, as seen in Fig.~\ref{fig:comp-ang} (left).
These advantages come at the cost of a moderately higher computation time, attributable to the dense connectivity of the global strain variable $\bm{q}$ within the factor graph.

\section{Experiments}
\label{sec:experiments}

\subsection{Experimental Setup}
\label{sec:exp-setup}
\begin{figure}[b]
\centerline{\includegraphics[width=\linewidth]{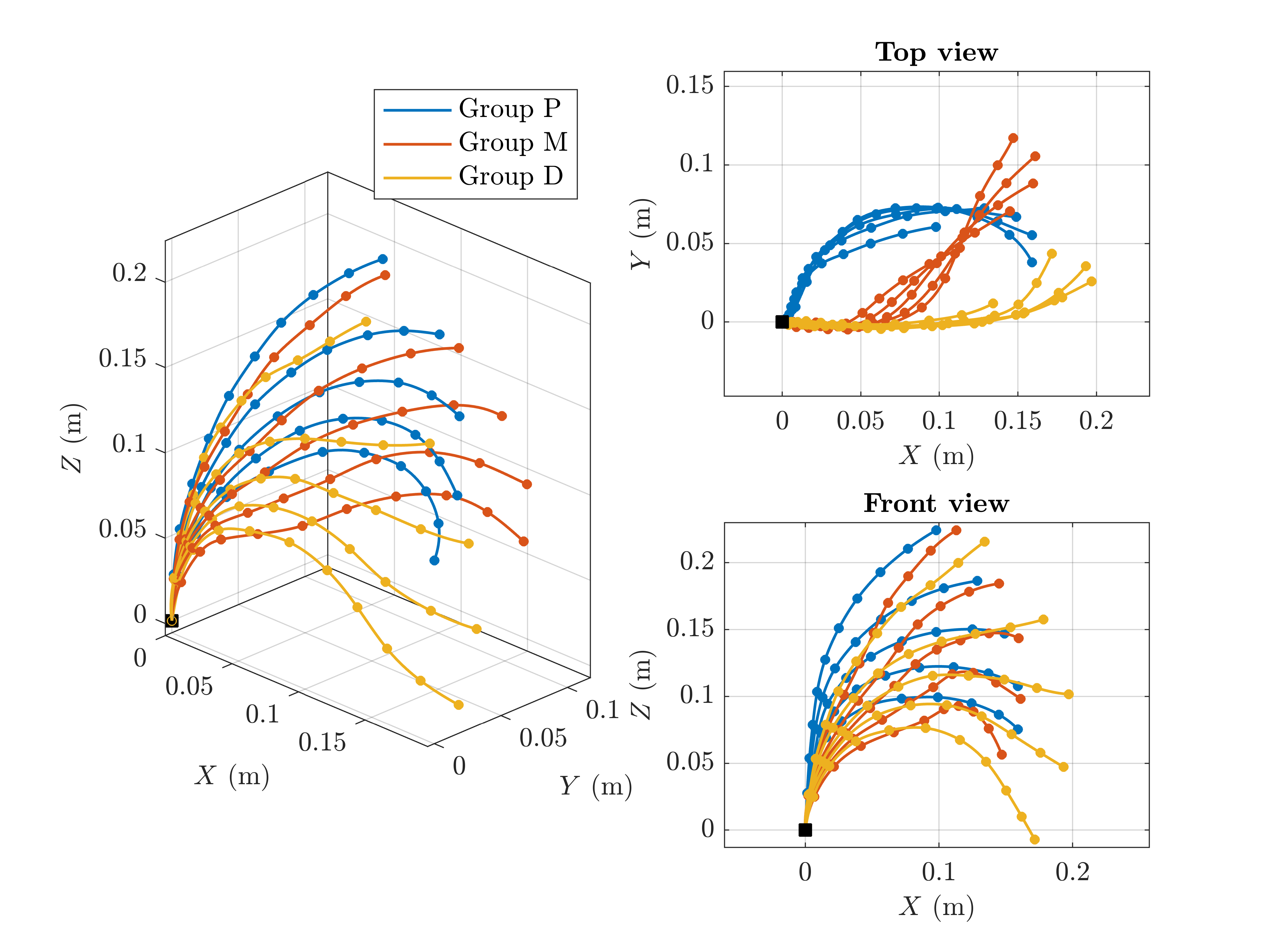}}
\caption{Diagram of the 15 experimental backbone configurations.
Markers $\bullet$ denote the backbone centerline positions obtained by triangulating the center of the spacer disks, as explained in Sec.~\ref{sec:exp-setup}.
The continuous backbone lines were obtained in post-processing by interpolating the disk locations with cubic polynomials.}
\label{fig:exp-shapes}
\end{figure}

The proposed framework was validated on data generated using a reconfigurable tendon-driven manipulator inspired by \cite{dash2026actuation}. 
For this version of the manipulator, smaller spacer disks were implemented which can be manually rotated about the backbone axis thus locally changing the tendon routing. 
\begin{figure}[t]
    \centering
    \subfloat[\label{fig:uiuc-scheme}]{%
        \includegraphics[width=0.39\linewidth]{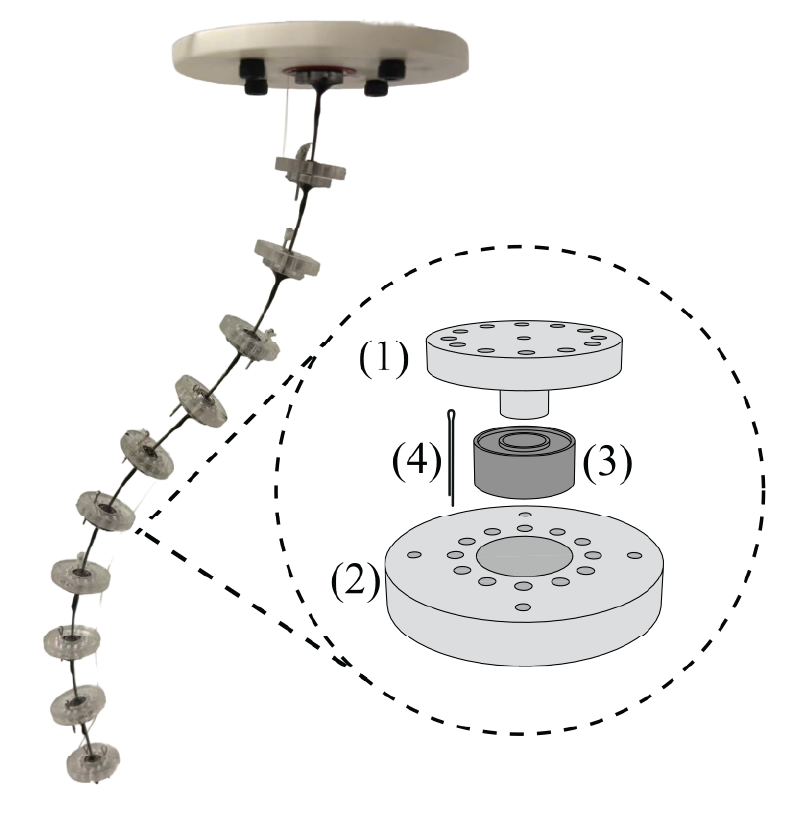}%
    }
    \hfill
    \subfloat[\label{fig:uiuc-sys}]{%
        \includegraphics[width=0.59\linewidth]{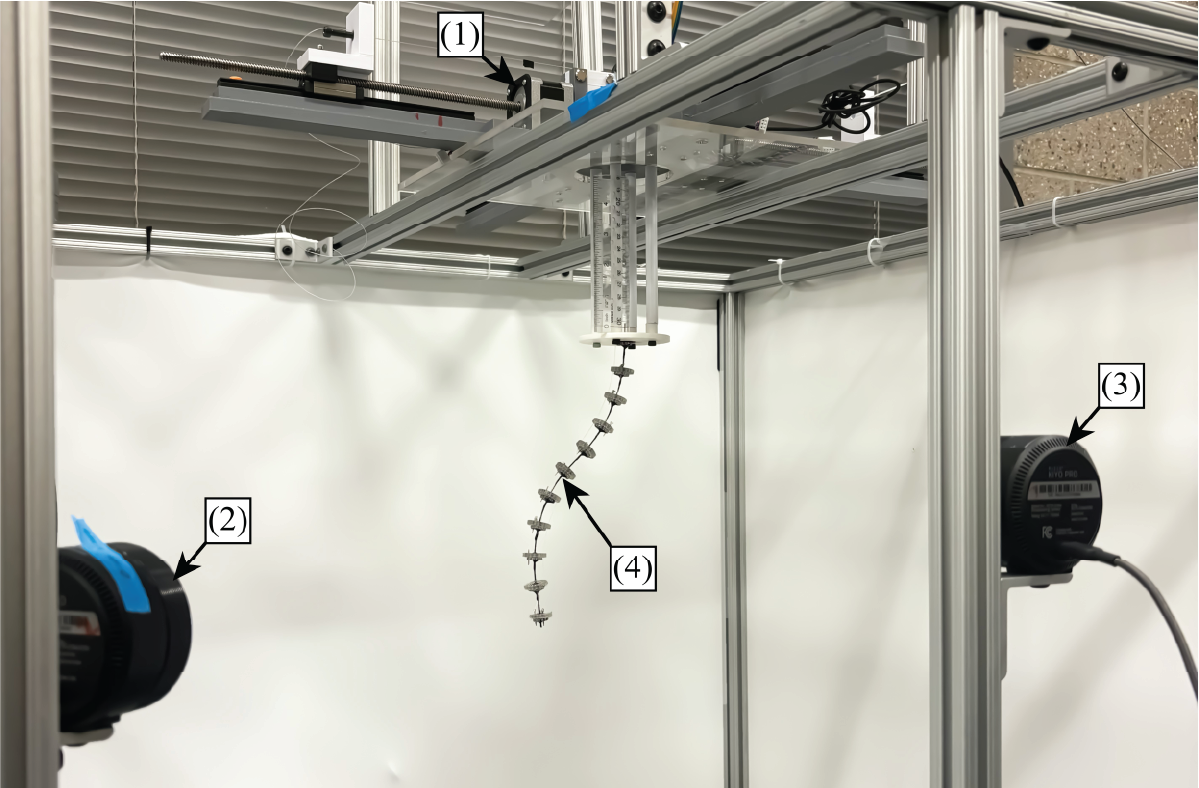}%
    }
    \caption{Fig.~\ref{fig:uiuc-scheme} depicts the manipulator prototype capable of changing the tendon routing by local rotation of intermediate spacer disks. 
    The exploded view of the spacer disk demonstrates the design: Part (1) is rigidly attached to the backbone, where Part (3) carries the tendon and is allowed to rotate freely on the bearing (2). 
    The disk (3) can be rotated and locked at discrete orientations using the cotter pin (4). 
    The experimental setup is shown in Fig.~\ref{fig:uiuc-sys}, where the tendons are actuated by the stepper motor (1), and the deformed profile is reconstructed using images from Cameras (2) and (3).}
    \label{fig:uiuc}
\end{figure}
As demonstrated in Fig.~\ref{fig:uiuc-scheme}, the manipulator achieves deformation through tendon actuation as well as local disk rotations. 
First, the disks are rotated and locked at specific positions, and then the tendon(s) are pulled to achieve the deformations.  
The deformed profile of the manipulator is captured using two RGB cameras (Razer Kiyo Pro - 1920 x 1080 at 60 fps, Sony STARVIS CMOS sensor, adjustable FOV 80°/90°/103°) rigidly fixed at two positions on the enclosure (Fig.~\ref{fig:uiuc-sys}). 
In this work, the manipulator centerline was reconstructed by triangulating the centers of 11 disks observed in both camera views. 
The two-dimensional disk centers were manually annotated from static images captured by each camera. 
For the validation purpose, 15 different experimental cases were evaluated, corresponding to three different groups of disks being rotated, denoted by letter P (disks 0-2), M (4-6), and D (8-10), for tendon displacements in steps of 5 mm ranging from 10 to 30 mm; the resulting shapes are reported in Fig.~\ref{fig:exp-shapes}.

\subsection{Hyperparameters and Measurement Scenarios}
\label{sec:exp-results}
In our experiments, the estimation factor graphs were instantiated with $N=11$ pose nodes, where the first is anchored to the manipulator base, and the remaining ten are placed at the spacer disk locations along the backbone.
For the proposed method, cubic B-splines were adopted as strain bases, consistently with Sec.~\ref{sec:simul}, with $n_{k_x}=n_{k_y}=6$ and $n_{p_x}=n_{p_y}=n_{p_z}=4$.
No DOFs were assigned to the twist component $k_z$, as it is unobservable under the position-only measurement setup considered here.
All remaining hyperparameters were kept identical to those in Sec.~\ref{sec:simul} for both the proposed and baseline methods.

Two measurement scenarios were considered, hereby denoted as E1 and E2.
In E1, position measurements were assigned to pose nodes 2, 4, 6, 8, and 10, yielding a uniform, distributed configuration expected to produce accurate shape estimates.
In E2, triangulated positions were associated only to nodes 6, 8, and 10, resulting in a sparser measurement setup in which the shape estimate is increasingly governed by the prior and by the dependency structure among estimation variables, thereby allowing a more demanding assessment of the proposed formulation.
Note that, although only a subset of the available triangulated backbone points was actively used in the graph, error statistics were computed by comparing the estimated position at all 11 nodes against the full triangulation result displayed in Fig.~\ref{fig:exp-shapes}, so as to evaluate each estimator's ability to infer the backbone shape in unobserved regions of the domain.
The overall accuracy and computational performance are reported in Table~\ref{tab:exp-stats}.

\begin{table}[b]
\caption{\label{tab:exp-stats}
Position error statistics and computational performance in scenarios E1 (position measurements at nodes 2, 4, 6, 8, 10) and E2 (position measurements at nodes 6, 8, 10).
}
\centering
\footnotesize
\setlength{\tabcolsep}{3pt}
\renewcommand{\arraystretch}{1.2}
\begin{tabular}{@{} l l l c c | c c @{}}
\hline
\hline
\textbf{Group} &
\textbf{ID} &
\textbf{Method} &
\begin{tabular}{@{}c@{}}\textbf{Pos.} \\ \textbf{Error}, mm\end{tabular} &
\begin{tabular}{@{}c@{}}\textbf{Max. Pos.} \\ \textbf{Error}, mm\end{tabular} &
\begin{tabular}{@{}c@{}}\textbf{Total} \\ \textbf{Time}, ms\end{tabular} &
\textbf{Iterations} \\
\hline
\multirow{4}{*}{P} &
\multirow{2}{*}{E1} & Proposed & \textbf{0.62$\pm$0.24} & \textbf{3.35} & 9.51$\pm$5.34 & 26.2$\pm$12.59 \\
 & & Baseline & 0.8$\pm$0.19 & 3.54 & \textbf{4.37$\pm$1.92} & \textbf{13$\pm$7.16} \\
\cline{2-7}
 & \multirow{2}{*}{E2} & Proposed & \textbf{3.26$\pm$0.71} & \textbf{15.28} & 5.82$\pm$1.72 & 14.4$\pm$4.59 \\
 & & Baseline & 4.83$\pm$1.13 & 21.38 & \textbf{4.11$\pm$2.69} & \textbf{10.2$\pm$4.96} \\
\hline
\multirow{4}{*}{M} &
\multirow{2}{*}{E1} & Proposed & \textbf{0.76$\pm$0.23} & 3.71 & 6.28$\pm$3.61 & 19.8$\pm$10.96 \\
 & & Baseline & 1.17$\pm$0.27 & \textbf{3.18} & \textbf{5.1$\pm$0.48} & \textbf{16.8$\pm$4.53} \\
\cline{2-7}
 & \multirow{2}{*}{E2} & Proposed & \textbf{2.49$\pm$1.41} & \textbf{14.21} & \textbf{3.98$\pm$0.94} & \textbf{11.4$\pm$1.62} \\
 & & Baseline & 7.6$\pm$2.59 & 36.59 & 6.11$\pm$1.55 & 16$\pm$7.82 \\
\hline
\multirow{4}{*}{D} &
\multirow{2}{*}{E1} & Proposed & \textbf{0.54$\pm$0.12} & 2.81 & 4.12$\pm$1.78 & 15.8$\pm$4.79 \\
 & & Baseline & 0.73$\pm$0.1 & \textbf{1.9} & \textbf{3.7$\pm$0.83} & \textbf{11.4$\pm$2.15} \\
\cline{2-7}
 & \multirow{2}{*}{E2} & Proposed & \textbf{1.61$\pm$0.34} & \textbf{6.14} & \textbf{5.16$\pm$2.06} & \textbf{17.2$\pm$6.18} \\
 & & Baseline & 4.76$\pm$1.98 & 20.24 & 10.36$\pm$8.45 & 29$\pm$23.05 \\
\hline
\hline
\end{tabular}
\end{table}

\subsection{Discussion}
\begin{figure}[t]
    \centerline{\includegraphics[width=\linewidth]{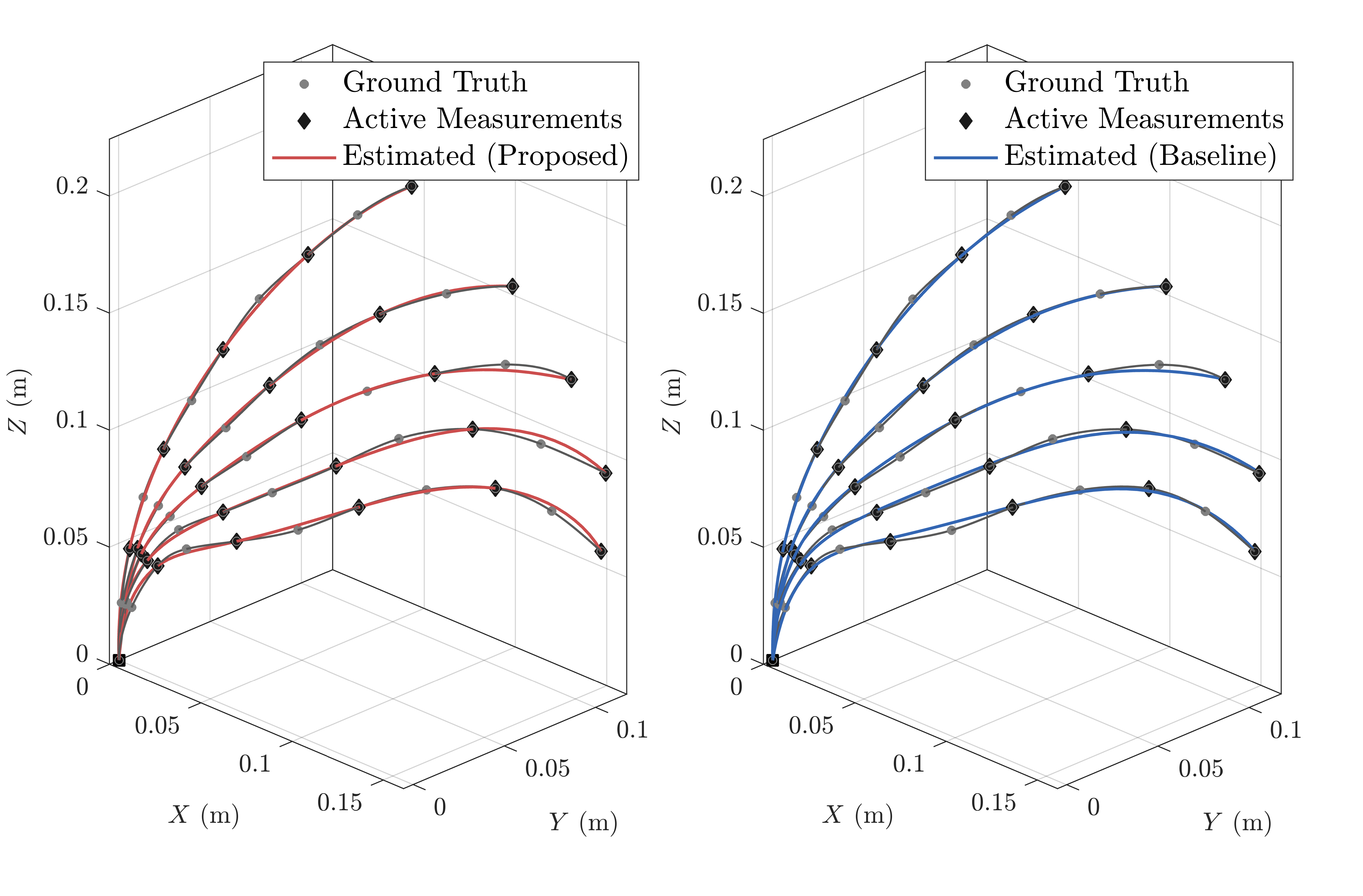}}
    \caption{Proposed (left) vs baseline (right) estimation result for samples M in scenario E1.}
    \label{fig:shapes-E1}
\end{figure}

\begin{figure}[t]
    \centerline{\includegraphics[width=\linewidth]{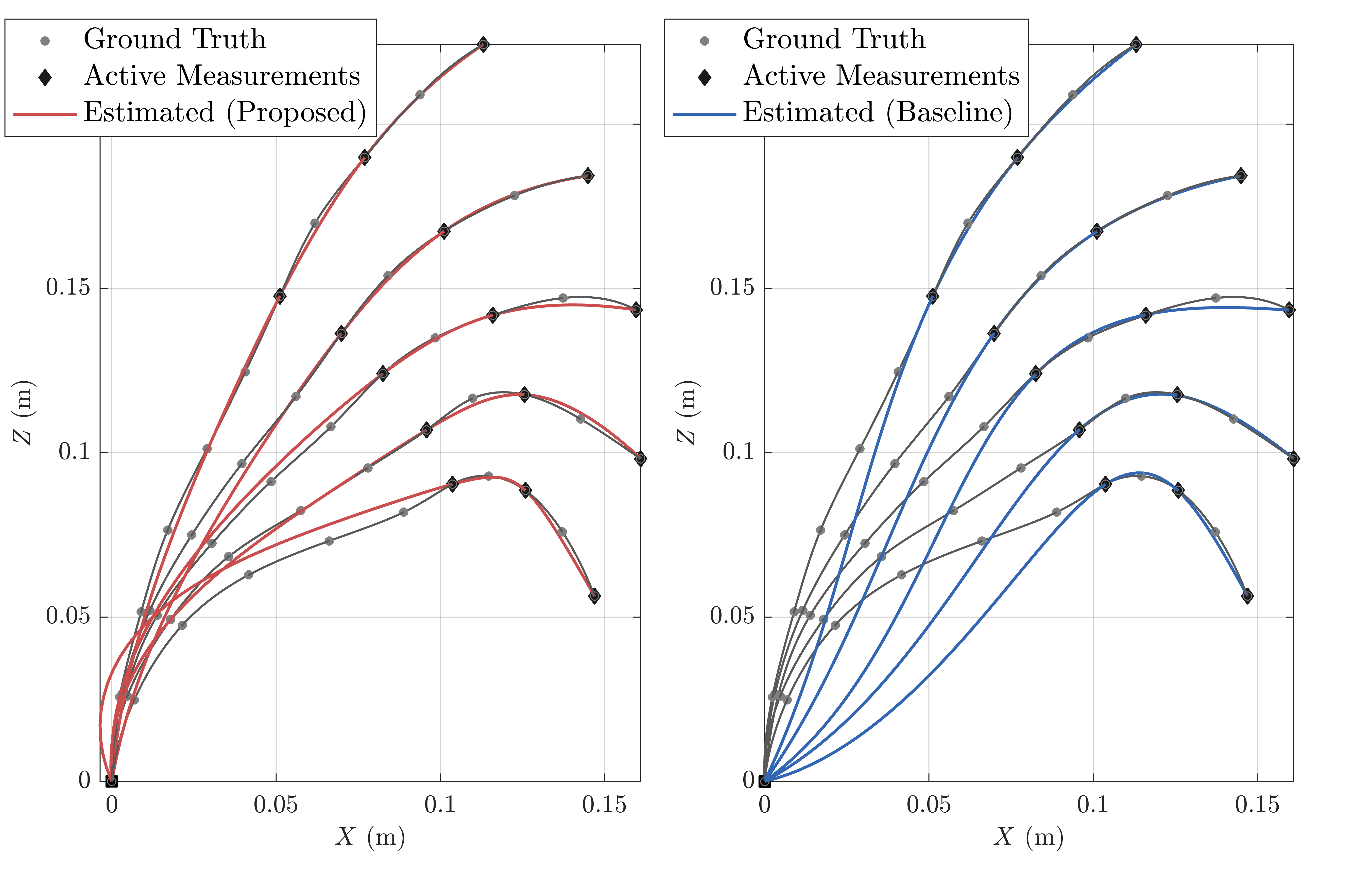}}
    \caption{Proposed (left) vs baseline (right) estimation result for samples M in scenario E2 (front view).}
    \label{fig:exp-E2-567-front}
\end{figure}

As expected, the best reconstruction accuracy is achieved under the distributed measurements of E1, as visible in Table~\ref{tab:exp-stats} and Fig.~\ref{fig:shapes-E1}.
In this scenario, the accuracy gap between the proposed method and the baseline is minimal. 
Notably, the baseline exhibits better computational efficiency owing to its fully sparse factor graph structure, which becomes increasingly beneficial as the number of measurement-related constraints grows.
The most pronounced performance disparity emerges in scenario E2, particularly for the configurations in group M.
This accuracy gap is clarified by Fig.~\ref{fig:exp-E2-567-front}: in the proximal region of the robot, where no measurements are available, the baseline reverts to its constant strain prior, introducing substantial reconstruction errors.

\section{Conclusion}
This letter presents a kinematics-based method for simultaneously reconstructing the pose and strain field of a continuum manipulator. 
The proposed approach formulates the estimation problem as a factor graph that encodes the probabilistic dependencies between variables and measurements, and leverages the GVS framework to approximate the strain field through a low-dimensional parameterization. 
Central to the formulation is a novel Magnus factor that acts as a strong kinematic prior, encoding the exact relationship between consecutive backbone pose nodes and the underlying strain field via the Magnus expansion. 
Compared to existing alternatives, the proposed approach allows the approximation order to be tailored to the available information, while the use of locally supported basis functions preserves numerical efficiency. 
Moreover, the Magnus factors capture higher-order geometric coupling between strain components, preserving estimation accuracy even in the presence of sparse or partial measurements. 
The method was evaluated in simulations and experiments, and an explicit comparison with a GP regression baseline demonstrated improved accuracy and more robust performance across several scenarios. 
Future work will extend the current framework to also reconstruct generalized system velocities and evaluate its suitability for model-based control.

\bibliographystyle{IEEEtran}
\bibliography{references}

\end{document}